\documentclass{article}
\usepackage{amsmath,graphicx,mlspconf}
\usepackage{xurl}

\usepackage{xcolor}

\usepackage{amssymb}
\usepackage{bm}
\usepackage{mathtools}
\usepackage{booktabs}
\usepackage{multirow}
\usepackage{subcaption}

\usepackage[breaklinks=true]{hyperref}

\DeclareMathOperator*{\argmin}{argmin}

\usepackage{pifont}
\newcommand{\cmark}{\ding{51}}
\newcommand{\xmark}{\ding{55}}

\toappear{2026 IEEE International Workshop on Machine Learning for Signal Processing, Sep.\ 28-- Oct.\ 1, 2026, Atlanta, USA}

\title{A COMPREHENSIVE BENCHMARK OF SOURCE-FREE UNIVERSAL DOMAIN ADAPTATION ON TIME SERIES REPRESENTATIONS}
\name{%
   Romain Mussard$^{\star}$
    \qquad Fannia  Pacheco$^{\star}$
    \qquad Maxime Berar$^{\star}$
    \qquad Paul Honeine$^{\star}$
    \qquad Gilles Gasso$^{\star}$\thanks{$^\star$ The authors acknowledge the support of the French Agence Nationale de la Recherche (ANR), under grant ANR-23-CE23-0004 (project ODD)}%
}
\address{%
   $^{\star}$ Univ Rouen Normandie, INSA Rouen Normandie, Université Le Havre Normandie, Normandie Univ,\\ LITIS UR 4108, F-76000 Rouen, France
}

\begin{document}

\maketitle

\begin{abstract}

Source-Free Universal Domain Adaptation (SF-UniDA) extends Universal Domain Adaptation by removing access to source data at adaptation time while still handling label-set mismatches between domains. Despite growing interest in this setting for image data, no benchmark exists for time series, which are more challenging. We present the first SF-UniDA benchmark on time series. In addition, we provide the first study of pretrained foundation models as feature extractors for time series domain adaptation. In this context, we identify a critical and previously underexplored limitation of all existing SF-UniDA methods: the inference threshold for unknown-sample rejection is highly sensitive. We address this by proposing a plug-in auto-thresholding module that can be integrated into any SF-UniDA method. Experiments on three well-known time series datasets confirm the suitability of this module. They also highlight that foundation models do not systematically outperform classical backbones and that SF-UniDA tailored for time series is yet to be developed.
\end{abstract}

\newcommand{\cem}[1]{\textcolor{blue}{cem: #1}}
%
\section{Introduction} 
\label{sec:intro}

Source-Free Universal Domain Adaptation (SF-UniDA) consists of aligning a model pretrained on a source distribution to an unlabeled target distribution. Its main objectives are common classes alignment between the two distributions and detection of target-only (\emph{unknown}) classes. This makes it applicable when source data cannot be shared due to privacy constraints \cite{UMAD}. When source data is available, one can instead resort to Universal Domain Adaptation (UniDA) \cite{UAN} techniques.

UniDA and SF-UniDA have been extensively studied for image datasets \cite{UMAD,GLC,GLC++,LEAD}, while time series datasets remain significantly understudied. Two benchmarks have evaluated domain adaptation for time series datasets: ADATIME \cite{ADATIME} addresses unsupervised domain adaptation (where both domains share the same label set), and UniDABench \cite{UniDABench} evaluates the impact of time-series-specific backbone architectures for the UniDA setting.
However, neither considers the SF-UniDA setting nor the use of foundation models as feature extractors. Our benchmark addresses both gaps within a unified framework for SF-UniDA.

SF-UniDA poses distinct challenges. For instance, without access to source data, adaptation must rely entirely on the source pretrained model, 
making the representational quality of its backbone all the more critical. The emergence of large pretrained foundation models for time series \cite{MOMENT,mantis,chronos} raises a question that no prior work has addressed: can their embeddings serve as effective feature representations for domain adaptation?
In addition, most SF-UniDA methods rely on a fixed and manually selected threshold to distinguish between common samples and unknown samples \cite{GLC, GLC++, LEAD}. Although SF-UniDA applications in computer vision demonstrate high robustness over threshold selection, a recent study cast doubt on such robustness over time series \cite{UniJDOT} in UniDA. This raises questions about the validity of manually selected thresholds in SF-UniDA.

This paper makes three contributions.  \textbf{(1)} We present the \emph{first benchmark for SF-UniDA on time series data}, contextualizing the foundation model results against four time-series-specific backbone architectures (CNN, TFE \cite{UniJDOT}, TSLANet \cite{TSLANet}, and S3 \cite{S3Layer}).  \textbf{(2)} We conduct the \emph{first study of pretrained foundation models for time series domain adaptation}, comparing MOMENT \cite{MOMENT}, Mantis \cite{mantis}, and Chronos \cite{chronos}.
\textbf{(3)} We identify and analyze \emph{threshold hypersensitivity} as a key limitation of all existing SF-UniDA methods on time series, and propose a plug-in auto-thresholding that resolves it without requiring labeled target data. 

\begin{figure*}[t]
\centering
\includegraphics[width=1.0\linewidth]{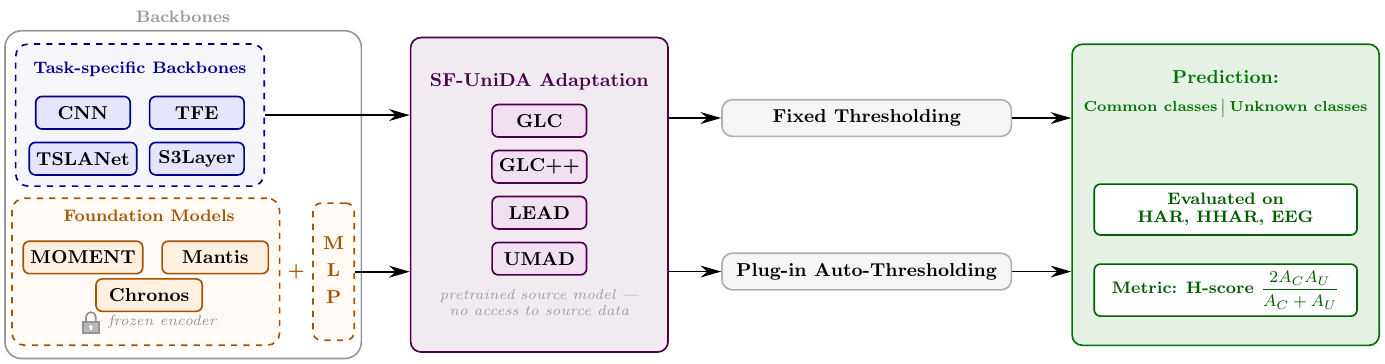}
\caption{Overview of the SF-UniDA benchmark for time series. Task-specific backbone architectures and pretrained foundation models are evaluated as feature extractors under the SF-UniDA methods. Both fixed thresholding and auto-thresholding are applied to each SF-UniDA method. The performance is measured with H-score over 3 time series datasets. }
\label{fig:benchmark_overview}
\end{figure*}

\section{Problem Formulation}
\label{sec:problem}

We consider a pretrained model $h = g \circ \phi$, with feature extractor $\phi: \mathcal{X} \to \mathcal{Z}$ and classifier $g: \mathcal{Z} \to \mathcal{Y}$, trained on a labeled source domain $\mathfrak{D}^s = \{(x_i^s, y_i^s)\}_{i=1}^{n_s}$ that is no longer accessible at adaptation time. The goal is to adapt $h$ to an unlabeled target domain $\mathfrak{D}^t = \{x_i^t\}_{i=1}^{n_t}$, drawn from a distribution $\mathcal{P}^t(x,y) \neq \mathcal{P}^s(x,y)$. We assume covariate shift: the marginal distributions differ while the labeling rule is preserved, i.e., $\mathcal{P}^s(x) \neq \mathcal{P}^t(x)$ and $\mathcal{P}^s(y|x) = \mathcal{P}^t(y|x)$.

Beyond distributional shift, the source and target label sets might not coincide \cite{UAN}. They decompose into three disjoint subsets: the \textbf{common label set} $\mathcal{Y} = \mathcal{Y}^s \cap \mathcal{Y}^t$ of classes shared by both domains, the \textbf{private source label set} $\overline{\mathcal{Y}}^s = \mathcal{Y}^s \setminus \mathcal{Y}$ of source-only classes absent from the target, and the \textbf{private target label set} $\overline{\mathcal{Y}}^t = \mathcal{Y}^t \setminus \mathcal{Y}$ of target-only \emph{unknown} classes unseen during training. We focus on the open-partial DA setting where both $\overline{\mathcal{Y}}^s \neq \emptyset$ and $\overline{\mathcal{Y}}^t \neq \emptyset$, the most complete and challenging scenario. This defines two key tasks: (i) aligning representations for common samples in $\mathcal{Y}$ across domains, and (ii) detecting target samples from $\overline{\mathcal{Y}}^t$ and assigning them to an \emph{unknown} category.  This setting, referred to as \textbf{Source-Free Universal Domain Adaptation} (SF-UniDA), is practically important when sharing source data is prohibited by privacy or governance constraints.
Multiple methods have been proposed for SF-UniDA in the image domain \cite{GLC,GLC++,LEAD,UMAD}, but none have been studied for time series applications. In addition, most of these methods rely on a fixed inference threshold to reject unknown samples, which have proven robust in the image domain, yet this robustness remains to be investigated for time series.


\section{Benchmark Design}
\label{sec:benchmark}

This benchmark applies all recent SF-UniDA methods to time series datasets and evaluates them across a diverse set of task-specific backbones and pretrained foundation models. We also investigate the robustness of the fixed inference threshold shared by all these methods in the time series setting, contrasting it against an auto-thresholding approach. The complete benchmark is illustrated in Fig.~\ref{fig:benchmark_overview}. Section~\ref{sec:backbones} presents the backbones and foundation models, Section~\ref{sec:models_datasets} describes the SF-UniDA methods, and Section~\ref{sec:threshold} motivates and details the proposed auto-thresholding module. Our code is available at \url{https://github.com/RomainMsrd/SF-UniDABench}.


\begin{table*}[ht]
\caption{Comparison of MOMENT, Mantis, and Chronos time series foundation models. General-purpose tasks include classification, forecasting, anomaly detection, and imputation.}
\label{tab:foundation_models}
\centering
\begin{tabular}{p{2.2cm}p{1cm}llp{2.5cm}p{3.5cm}p{2.4cm}}
\toprule
\textbf{Model} & \textbf{Type} & \textbf{Architecture} & \textbf{Param.} & \textbf{Input} & \textbf{Pre-training objective} & \textbf{Task} \\
\midrule
MOMENT \cite{MOMENT} & Encoder & T5 & 40M & Patches & Reconstruction & General-purpose \\
Mantis \cite{mantis} & Encoder & ViT & 8M & Tokens & Contrastive learning& Classification \\
Chronos \cite{chronos} & Seq2Seq & T5 and LLM & 8M  & Scalar quantized & Autoregression & Forecasting \\
\bottomrule
\end{tabular}

\end{table*}

\begin{table*}[t]
\centering
\caption{Comparison of SF-UniDA methods. Threshold-free refers to inference only. ``Weak'' source-free indicates UMAD requires a dual-head architecture and auxiliary orthogonal loss during source pre-training.}
\label{tab:sfunida_comparison}
\begin{tabular}{lllccc}
\toprule
\textbf{Method} & \textbf{Key idea} & \textbf{Unknown detection} & \textbf{Source-free} & \textbf{Threshold-free (Inference)} \\
\midrule
UMAD \cite{UMAD}   & Dual-head classifier consistency      &  Consistency + MixUP    & Weak & \cmark \\
GLC \cite{GLC}     & Global One Vs All + local k-NN     & One Vs All clustering & Full & \xmark \\
GLC++ \cite{GLC++} & GLC + contrastive loss    & One Vs All clustering & Full & \xmark \\
LEAD \cite{LEAD}  & Orthogonal feature decomposition   & Gaussian Mixture Model  & Full & \xmark \\
\bottomrule
\end{tabular}
\end{table*}

\subsection{Backbones \& Foundation Models}
\label{sec:backbones}

\textbf{Comparison backbones.} We use four time-series-specific architectures as baselines. \textbf{1D-CNN} (three-block convolutional network) serves as the reference baseline. \textbf{TFE} \cite{UniJDOT} concatenates a Fourier-based encoder with a parallel CNN branch to jointly capture spectral and temporal features. \textbf{TSLANet} \cite{TSLANet} combines learnable spectral filtering with gated convolutional mixing. \textbf{S3Layer} \cite{S3Layer} segments the series into patches, reorders them via a learned priority vector, and stitches them back with a residual connection. 
\textbf{Foundation models.} Three foundation models are evaluated in this work. They have been trained to solve different tasks and have been chosen for their relatively small number of parameters. These foundation models are summarized in Table~\ref{tab:foundation_models} and span a range of pretraining objectives and tasks.  

\textbf{MOMENT} \cite{MOMENT} is a general-purpose T5-style encoder pretrained via masked patch reconstruction on a large collection of public time series spanning multiple tasks (forecasting, classification, anomaly detection, and imputation). Its multi-task pretraining distinguishes it from Mantis, which is designed specifically for classification.

\textbf{Mantis} \cite{mantis} is a ViT-based encoder designed specifically for time series classification. Its Token Generator Unit fuses three complementary views of each series (instance-normalized patches, first-order differences, and global statistics) and the encoder is pretrained via a contrastive objective, making it the most naturally suited model for the classification task at hand.

\textbf{Chronos} \cite{chronos} adapts the T5 language model to time series forecasting by quantizing observations into discrete scalar tokens and training autoregressively on a large corpus of real and synthetic time series. Despite being designed for forecasting, its encoder embeddings are used here as a feature extractor, representing the extreme case of a model pretrained on a fundamentally different task than classification.

\subsection{SF-UniDA Methods}
\label{sec:models_datasets}

The considered SF-UniDA methods are summarized in Table ~\ref{tab:sfunida_comparison} and further described below.

\textbf{UMAD} \cite{UMAD} trains a source model with a two-head classifier regularized by an orthogonality constraint, so that the two heads learn complementary decision boundaries. At adaptation and inference time, unknown samples are detected with an information-consistency score, defined as the inner product between the two heads' softmax outputs.

\textbf{GLC} \cite{GLC} proposes a Global and Local Clustering approach. It combines an adaptive one-vs-all global clustering algorithm to distinguish common and unknown target classes, with a local $k$-NN clustering strategy to mitigate negative transfer from unknown samples.

\textbf{GLC++} \cite{GLC++} extends GLC by integrating a contrastive affinity learning strategy to overcome the limitation of uniform treatment of unknown data inherent to closed-set source architectures, enabling finer discrimination among distinct unknown categories.

\textbf{LEAD} \cite{LEAD} proposes a LEArning Decomposition framework that decouples target features into source-common and source-unknown components via orthogonal decomposition. Unknown samples are then identified by thresholding the norm of each sample's source-unknown component, avoiding time-consuming iterative clustering.

All methods except UMAD rely on a manually selected threshold to reject unknown samples at inference. UMAD instead estimates its threshold from MixUp-interpolated target samples acting as synthetic unknowns, but requires a specific dual-head architecture and source pre-training objective, unlike other methods, which can adapt any off-the-shelf classifier. Therefore, UMAD is included for comparison while noting its slightly relaxed SF-UniDA assumption.


\begin{figure}
    \centering
    \includegraphics[width=1.0\linewidth]{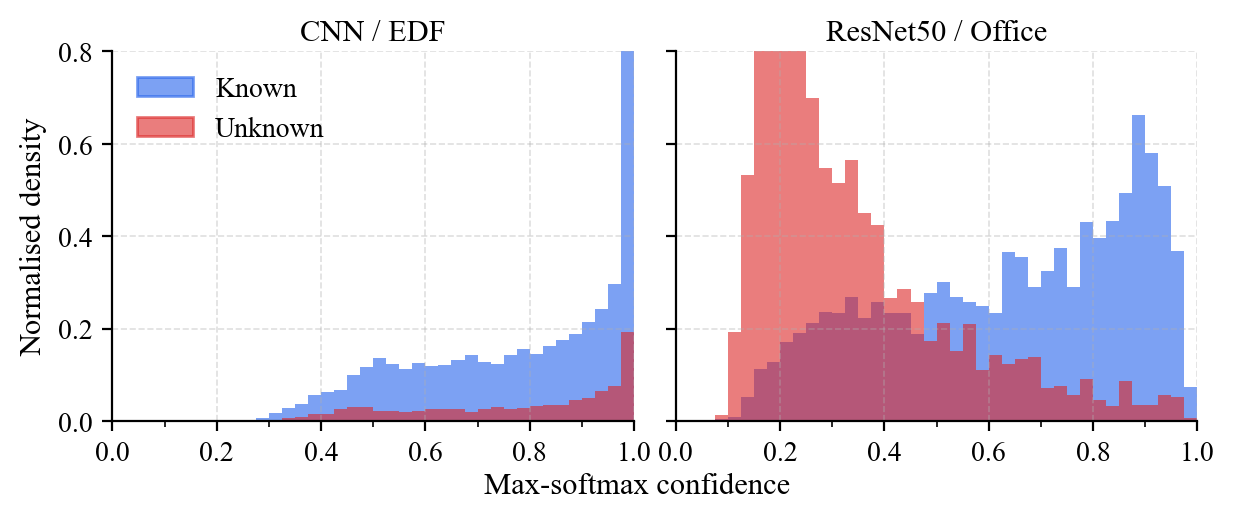}
    \caption{Source model confidence distributions on target data (no adaptation) for known and unknown samples. Left: CNN / EDF (Time Series). Right: ResNet50 / Office-31 (Images).}
    \label{fig:confidence_comparison}
\end{figure}

\subsection{Threshold Sensitivity in SF-UniDA}
\label{sec:threshold}

Strictly source-free SF-UniDA methods reject unknown target samples by thresholding the maximum softmax probability with a fixed manually selected value $\tau$. In computer vision, sensitivity analyses report marginal H-score variations across broad ranges of $\tau$, making this choice appear relatively harmless \cite{GLC}. We argue this assumption should be revisited for time series, as dynamic auto-thresholding has already shown promise for unknown-class detection in this setting \cite{UniJDOT}, suggesting that fixed thresholds may be less reliable. 

We hypothesize that common and unknown confidence distributions are harder to separate due to noisier and less structured feature spaces producing less calibrated softmax outputs (see Fig.~\ref{fig:confidence_comparison}). Under this hypothesis, the H-score can become highly sensitive to $\tau$, with small threshold changes causing large performance swings and narrow performance peaks that exhaustive grid search may miss. We therefore compare each SF-UniDA method with and without a drop-in auto-thresholding block replacing the fixed threshold, which additionally reduces the hyperparameter search space.

\begin{figure*}[t]
    \centering
    \includegraphics[width=0.49\linewidth]{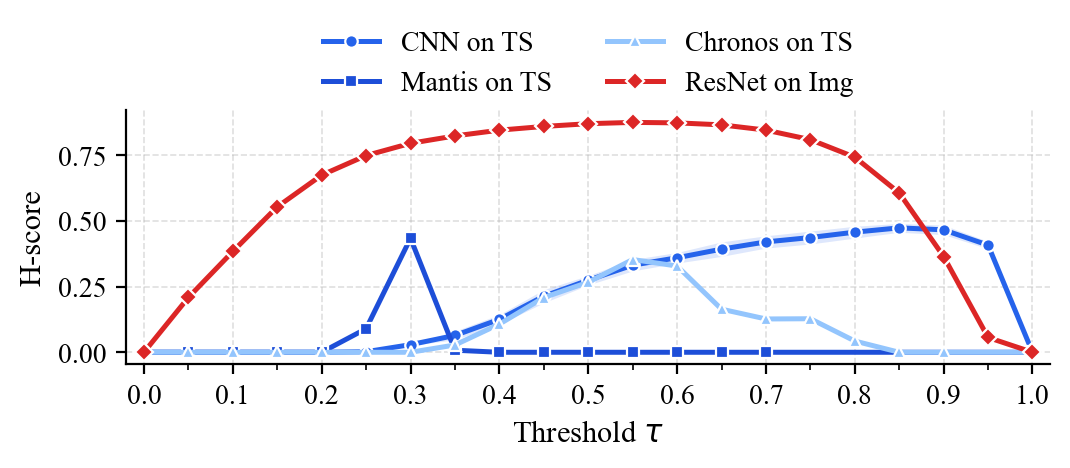}%
    \includegraphics[width=0.49\linewidth]{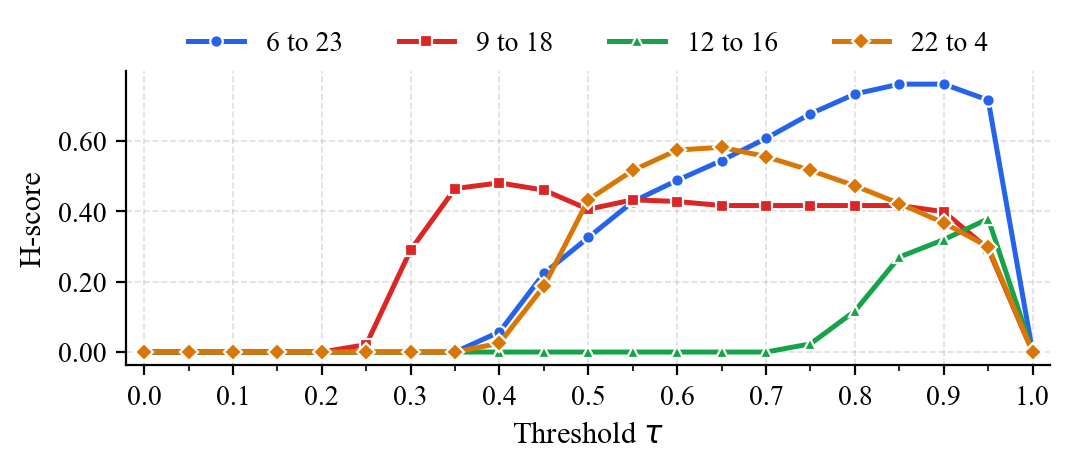}
    \caption{\textbf{Threshold sensitivity.} \textit{Left:} Mean H-score as a function of $\tau$ for GLC on time series (HAR) and image (Office) data with a CNN backbone and two foundation models, showing that the optimal threshold range is far narrower for time series. \textit{Right:} Per-scenario H-score as a function of $\tau$ for GLC with a CNN backbone on HAR, illustrating that the optimal threshold also varies across adaptation scenarios.}
    \label{fig4:tau_sensitivity_image}
\end{figure*}

\textbf{Plug-in auto-thresholding module.}
To address this limitation, we propose a drop-in module applicable to any SF-UniDA method that replaces the fixed threshold with one estimated automatically from the unlabeled target data.
After the adaptation step, the empirical distribution of classifier confidences $\{s_i\}_{i=1}^{|\mathcal{D}_{tr}^{t}|}$, where $s_i = \max_c\, p_{i,c}^t$, is computed on the unlabeled training target set $\mathcal{D}_{tr}^{t}$. An unsupervised histogram-based thresholding criterion $\mathcal{F}$ is then applied to this distribution to yield the optimal threshold:
\begin{equation}
  \label{eq:autothresh}
  \tau^* = \argmin_\tau \mathcal{F}\!\left(\{s_i\}_{i=1}^{|\mathcal{D}_{tr}^{t}|}\right).
\end{equation}
Any criterion that partitions a one-dimensional distribution into two groups without requiring labels is a valid instantiation of $\mathcal{F}$. We consider three standard choices: \textbf{Otsu's criterion}~\cite{otsu} which minimizes intra-class variance, \textbf{Yen's criterion}~\cite{yen_thresh} which maximizes the entropic correlation of each partition, and \textbf{Li's criterion}~\cite{li_thr} which minimizes cross-entropy to group means. The resulting $\tau^*$ is computed once after adaptation and applied fixed at test time, adding negligible overhead, requiring no labels, and no modification to the adaptation procedure, making it a true plug-in compatible with any fully SF-UniDA method~\cite{UniJDOT}.

\section{Experiments}
\label{sec:experiments}

\subsection{Experimental Setup}

SF-UniDA methods are evaluated on the datasets used in \cite{UniDABench}: 
\textbf{HAR} and \textbf{HHAR} are human activity recognition datasets with six activity classes (walking, sitting, biking, etc.) collected from body-worn sensors, both use 128-timestep windows and are multivariate (9 and 3 channels, respectively). \textbf{EDF} is a sleep stage classification dataset with five sleep stages from EEG recordings. We also consider the image dataset Office31 \cite{office31} for comparison with time series. Backbone hyperparameters are fixed across all experiments, foundation models are frozen, and a trainable single-hidden-layer MLP is appended on top to let the SF-UniDA method adjust the feature space without modifying the foundation model weights. The performance is measured by the H-score $(2A_C A_U)/(A_C + A_U)$ \cite{CMU}, where $A_C$ and $A_U$ denote accuracy on common and unknown target classes, respectively.

All SF-UniDA methods rely on their own set of hyperparameters, which generally include a fixed threshold $\tau$ during inference, whose selection deeply impacts performance. Following \cite{UniDABench}, hyperparameters are selected by Bayesian optimization over $N_r$ trials. Given that domain adaptation datasets are structured as a collection of $d$ domains $\mathbf{D} = \{\mathcal{D}^1, \ldots, \mathcal{D}^d\}$, where any ordered pair $(\mathcal{D}^s, \mathcal{D}^t)$ with $s \neq t$ defines an adaptation scenario, the full set of scenarios $\mathcal{S}$ can be partitioned into a validation subset $\mathcal{S}_{val}$ of size $N_{val}$ and an evaluation subset $\mathcal{S}_{eval}$ of size $N_{eval}$ for final reporting. This scenario-level separation ensures that the H-score signal used for selection does not contaminate the final evaluation. We use 20 training epochs per model, $N_r = 20$ Bayesian hyperparameter trials, $N_{val} = 5$ validation scenarios for HAR and $N_{val} = 3$ for HHAR and EDF, with final results over $N_{eval} = 10$ test scenarios each averaged over 10 seeds. Note that this benchmark protocol requires labeled target validation scenarios.

\begin{table*}[t]
\centering
\caption{H-score (\%) with Auto-Thresholding (Yen's criterion) over 10 seeds. Parentheses show gain vs. fixed thresholding: ($\uparrow$) gain, ($\downarrow$) loss, ($=$) no change. Bold and italic indicate the best and second-best architectures for each method, respectively. UniJDOT in gray is a UniDA method for which backbone results marked with $^*$ are reported directly from \cite{UniDABench}.}
\resizebox{\linewidth}{!}{
\begin{tabular}{l l c c c c c c c}
\toprule
\multirow{3}{*}{\textbf{Datasets}} & \multirow{3}{*}{\textbf{Methods}} & \multicolumn{4}{c}{\textbf{Backbones}} & \multicolumn{3}{c}{\textbf{Foundation Models}}\\
\cmidrule(lr){3-6} \cmidrule(lr){7-9}
& & \textbf{CNN} & \textbf{TFE} & \textbf{S3} & \textbf{TSLANet} & \textbf{Mantis} & \textbf{Moment} & \textbf{Chronos}\\
\midrule 
\multirow{5}{*}{\textbf{HAR}} & \textcolor{gray!65!black}{\textbf{UniJDOT}} & \textcolor{gray!65!black}{61.0$^*$ (--)}& \textcolor{gray!65!black}{64.6$^*$ (--)}& \textcolor{gray!65!black}{54.1$^*$ (--)}& \textcolor{gray!65!black}{59.7$^*$ (--)}& \textcolor{gray!65!black}{\textbf{82.5 (--)}} & \textcolor{gray!65!black}{30.1 (--)} & \textcolor{gray!65!black}{\textit{72.2 (--)}} \\
 & \textbf{UMAD} & \textit{52.7 \textcolor{red}{($\downarrow$ 2.5)}} & \textbf{64.6 \textcolor{green!60!black}{($\uparrow$ 29.7)}} & 36.1 \textcolor{red}{($\downarrow$ 7.5)} & 43.1 \textcolor{green!60!black}{($\uparrow$ 11.5)} & 51.2 \textcolor{green!60!black}{($\uparrow$ 16.2)} & 24.2 \textcolor{green!60!black}{($\uparrow$ 11.4)} & 46.0 \textcolor{green!60!black}{($\uparrow$ 13.3)} \\
 & \textbf{GLC} & 46.2 \textcolor{green!60!black}{($\uparrow$ 4.2)} & 42.2 \textcolor{green!60!black}{($\uparrow$ 29.0)} & 52.3 \textcolor{green!60!black}{($\uparrow$ 11.2)} & \textbf{54.0 \textcolor{green!60!black}{($\uparrow$ 5.8)}} & \textit{53.4 \textcolor{green!60!black}{($\uparrow$ 15.8)}} & 16.7 \textcolor{green!60!black}{($\uparrow$ 16.6)} & 50.6 \textcolor{green!60!black}{($\uparrow$ 28.9)} \\
 & \textbf{GLC++} & 44.7 \textcolor{green!60!black}{($\uparrow$ 5.4)} & 36.4 \textcolor{green!60!black}{($\uparrow$ 20.8)} & 50.7 \textcolor{green!60!black}{($\uparrow$ 5.0)} & \textit{54.1 \textcolor{green!60!black}{($\uparrow$ 17.8)}} & \textbf{59.6 \textcolor{green!60!black}{($\uparrow$ 51.1)}} & 13.6 \textcolor{green!60!black}{($\uparrow$ 10.8)} & 50.9 \textcolor{green!60!black}{($\uparrow$ 33.8)} \\
 & \textbf{LEAD} & 43.8 \textcolor{green!60!black}{($\uparrow$ 7.2)} & 42.3 \textcolor{green!60!black}{($\uparrow$ 28.1)} & 50.4 \textcolor{green!60!black}{($\uparrow$ 4.4)} & \textit{53.1 \textcolor{green!60!black}{($\uparrow$ 15.5)}} & \textbf{57.8 \textcolor{green!60!black}{($\uparrow$ 13.3)}} & 13.7 \textcolor{green!60!black}{($\uparrow$ 11.5)} & 50.6 \textcolor{green!60!black}{($\uparrow$ 30.7)} \\
\midrule
\multirow{5}{*}{\textbf{HHAR}} & \textcolor{gray!65!black}{\textbf{UniJDOT}} & \textcolor{gray!65!black}{56.6$^*$ (--)}& \textcolor{gray!65!black}{\textbf{61.2}$^*$ (--)}& \textcolor{gray!65!black}{57.8$^*$ (--)}& \textcolor{gray!65!black}{55.7$^*$ (--)}& \textcolor{gray!65!black}{\textit{59.3 (--)}} & \textcolor{gray!65!black}{46.3 (--)} & \textcolor{gray!65!black}{57.6 (--)} \\
 & \textbf{UMAD} & \textbf{53.7 \textcolor{green!60!black}{($\uparrow$ 0.5)}} & 42.6 \textcolor{green!60!black}{($\uparrow$ 21.1)} & \textit{51.5 \textcolor{red}{($\downarrow$ 2.0)}} & 41.1 \textcolor{green!60!black}{($\uparrow$ 2.7)} & 28.6 \textcolor{green!60!black}{($\uparrow$ 4.5)} & 34.1 \textcolor{red}{($\downarrow$ 10.2)} & 40.2 \textcolor{green!60!black}{($\uparrow$ 1.6)} \\
 & \textbf{GLC} & 43.2 \textcolor{green!60!black}{($\uparrow$ 14.1)} & \textit{45.3 \textcolor{green!60!black}{($\uparrow$ 13.4)}} & \textbf{45.5 \textcolor{green!60!black}{($\uparrow$ 3.2)}} & 29.2 \textcolor{green!60!black}{($\uparrow$ 1.4)} & 39.4 \textcolor{green!60!black}{($\uparrow$ 27.1)} & 31.3 \textcolor{green!60!black}{($\uparrow$ 17.4)} & 31.6 \textcolor{red}{($\downarrow$ 3.8)} \\
 & \textbf{GLC++} & 37.3 \textcolor{green!60!black}{($\uparrow$ 10.3)} & \textit{44.9 \textcolor{green!60!black}{($\uparrow$ 13.2)}} & \textbf{47.7 \textcolor{green!60!black}{($\uparrow$ 7.9)}} & 29.6 \textcolor{green!60!black}{($\uparrow$ 2.7)} & 37.7 \textcolor{green!60!black}{($\uparrow$ 27.0)} & 26.8 \textcolor{green!60!black}{($\uparrow$ 18.7)} & 25.3 \textcolor{red}{($\downarrow$ 14.9)} \\
 & \textbf{LEAD} & 41.0 \textcolor{green!60!black}{($\uparrow$ 12.1)} & \textit{45.6 \textcolor{green!60!black}{($\uparrow$ 13.0)}} & \textbf{52.2 \textcolor{green!60!black}{($\uparrow$ 11.2)}} & 17.8 \textcolor{green!60!black}{($\uparrow$ 10.0)} & 40.2 \textcolor{green!60!black}{($\uparrow$ 22.9)} & 35.0 \textcolor{green!60!black}{($\uparrow$ 18.5)} & 29.4 \textcolor{green!60!black}{($\uparrow$ 25.3)} \\
\midrule
\multirow{5}{*}{\textbf{EDF}} & \textcolor{gray!65!black}{\textbf{UniJDOT}} & \textcolor{gray!65!black}{44.3$^*$ (--)}& \textcolor{gray!65!black}{\textbf{55.6}$^*$ (--)}& \textcolor{gray!65!black}{50.4$^*$ (--)}& \textcolor{gray!65!black}{\textit{55.0}$^*$ (--)}& \textcolor{gray!65!black}{42.4 (--)} & \textcolor{gray!65!black}{37.7 (--)} & \textcolor{gray!65!black}{38.1 (--)} \\
 & \textbf{UMAD} & 46.5 \textcolor{green!60!black}{($\uparrow$ 4.9)} & 32.9 \textcolor{red}{($\downarrow$ 1.4)} & \textit{48.1 \textcolor{green!60!black}{($\uparrow$ 2.5)}} & \textbf{49.3 \textcolor{green!60!black}{($\uparrow$ 7.7)}} & 29.3 \textcolor{green!60!black}{($\uparrow$ 10.9)} & 30.5 \textcolor{green!60!black}{($\uparrow$ 5.6)} & 34.8 \textcolor{green!60!black}{($\uparrow$ 4.9)} \\
 & \textbf{GLC} & \textbf{46.7 \textcolor{green!60!black}{($\uparrow$ 1.8)}} & 24.5 \textcolor{red}{($\downarrow$ 14.6)} & \textit{38.8 \textcolor{green!60!black}{($\uparrow$ 1.3)}} & 31.8 \textcolor{green!60!black}{($\uparrow$ 2.9)} & 30.9 \textcolor{green!60!black}{($\uparrow$ 10.7)} & 31.3 \textcolor{green!60!black}{($\uparrow$ 6.8)} & 30.4 \textcolor{green!60!black}{($\uparrow$ 2.9)} \\
 & \textbf{GLC++} & \textbf{49.5 \textcolor{green!60!black}{($\uparrow$ 1.6)}} & 31.3 \textcolor{red}{($\downarrow$ 10.8)} & \textit{43.8 \textcolor{green!60!black}{($\uparrow$ 1.7)}} & 32.6 \textcolor{green!60!black}{($\uparrow$ 2.9)} & 27.4 \textcolor{green!60!black}{($\uparrow$ 12.4)} & 28.3 \textcolor{green!60!black}{($\uparrow$ 2.5)} & 29.6 \textcolor{green!60!black}{($\uparrow$ 1.5)} \\
 & \textbf{LEAD} & \textbf{46.2 \textcolor{green!60!black}{($\uparrow$ 0.2)}} & 36.3 \textcolor{green!60!black}{($\uparrow$ 7.4)} & \textit{44.6 (=)} & 33.7 \textcolor{green!60!black}{($\uparrow$ 4.8)} & 28.9 \textcolor{green!60!black}{($\uparrow$ 10.2)} & 37.0 \textcolor{green!60!black}{($\uparrow$ 8.4)} & 27.2 \textcolor{green!60!black}{($\uparrow$ 3.1)} \\
\bottomrule
\end{tabular}
}
\label{tab:h_score_summary}
\end{table*}

\subsection{Threshold Sensitivity Analysis}
\label{sec:threshold_analysis}  

The Fig.~\ref{fig4:tau_sensitivity_image} (left) reveals that the optimal fixed threshold for time series occupies a much narrower range than for image data. This is especially pronounced for foundation models such as Mantis and Chronos, whose H-score collapses to nearly zero across most of the range and only spikes sharply. Even with a CNN backbone on time series, although the H-score increases more gradually from $\tau = 0.3$ to $\tau = 0.9$, it then drops abruptly to zero, with no flat plateau. This contrasts with GLC using a ResNet backbone on image data, which maintains a stable and high H-score over a wide range of thresholds before degrading near the extremes. The right panel further shows that the optimal threshold varies across adaptation scenarios, so a fixed threshold tuned at the dataset level may not generalize well within the same dataset.

\subsection{Benchmark Results}

Table~\ref{tab:h_score_summary} reports H-scores with gains over the best fixed thresholds found via hyperparameter search in parentheses. UMAD frequently leads among SF-UniDA methods as a weakly source-free method (i.e., requiring a specific architecture and source pretraining), while foundation models with a trainable MLP do not consistently exceed standard backbones: Mantis is competitive on HAR, whereas on HHAR and EDF, classical backbones generally match or outperform all foundation models. Moment's low H-scores for HAR are traceable to poor source model accuracy despite equivalent hyperparameter tuning rather than to the adaptation procedure. The top non-source-free method, UniJDOT~\cite{UniJDOT} outperforms SF-UniDA, confirming a persistent gap with UniDA.

Auto-thresholding generally improves performance across methods and backbones, with the largest gains for foundation models. On HAR, Mantis achieves the highest foundation-model H-scores, with gains exceeding 50 points for GLC++ and 10 points otherwise, while Chronos gains around 30 points for most methods, confirming that fixed thresholds severely underestimate foundation models. On HHAR and EDF, gains remain predominant, with isolated losses for Chronos on HHAR and TFE on EDF. The proposed plug-in also outperforms UMAD's own auto-thresholding on average.

As shown in Table~\ref{tab:thr_glc_har}, Yen’s criterion ranks first or second while Otsu’s and Li’s remain competitive, confirming that other criteria can be used with no or little performance loss. These results demonstrate that the proposed auto-thresholding module effectively addresses the critical threshold sensitivity issue in time-series SF-UniDA.

\begin{table}[t]
\centering
\caption{H-score (\%) over 10 seeds for \textbf{GLC} on \textbf{HAR} across backbones and thresholding methods}
\resizebox{\linewidth}{!}{
\begin{tabular}{l c c c c c}
\toprule
\multirow{3}{*}{\textbf{Threshold}} & \multicolumn{3}{c}{\textbf{Backbones}} & \multicolumn{2}{c}{\textbf{Foundation M.}} \\
\cmidrule(lr){2-4} \cmidrule(lr){5-6}
& \textbf{CNN} & \textbf{TFE} & \textbf{S3Layer} & \textbf{Mantis} & \textbf{Chronos}\\
\midrule
\textbf{yen} & \textbf{46.2} & \underline{42.2} & \underline{52.3} & \textbf{53.4} & \textbf{50.6} \\
\textbf{otsu} & \underline{44.5} & \textbf{42.9} & 49.8 & 45.7 & \underline{42.9} \\
\textbf{li} & 42.0 & 41.3 & \textbf{55.3} & \underline{48.0} & 40.3 \\
\bottomrule
\end{tabular}
}
\label{tab:thr_glc_har}
\end{table}

\section{Conclusion}
\label{sec:conclusion}

We presented the first SF-UniDA benchmark for time series, evaluating four backbones and four SF-UniDA methods across three datasets, as well as pretrained foundation models as feature extractors for time-series domain adaptation. Results show that S3Layer ranks among the top classical backbones, that SF-UniDA lags behind its UniDA counterpart, and that frozen foundation model features with an additional MLP do not systematically outperform task-specific backbones, although other fine-tuning strategies may yield different results. We further identified threshold hypersensitivity as a critical and underexplored limitation of existing SF-UniDA methods on time series, and proposed a label-free plug-in auto-thresholding module that consistently improves performance across all methods and backbones, with gains exceeding 50 H-score points for some foundation models. Finally, backbone selection currently has a larger impact on performance than the choice of SF-UniDA method, suggesting that developing time-series-specific SF-UniDA approaches is a promising direction for future work.




\bibliographystyle{IEEEbib}
\bibliography{refs}

\end{document}